\documentclass{article}
\usepackage[a4paper, total={6in, 8in}]{geometry}
\usepackage{moreverb}
\usepackage{multicol}
\usepackage{multirow}
\usepackage{subcaption}
\usepackage{hyperref}
\usepackage{todonotes}
\usepackage{xspace}
\usepackage{amsfonts}

\hypersetup{colorlinks,citecolor=black,filecolor=black,linkcolor=black, urlcolor=black}
\newcommand\BibTeX{{\rmfamily B\kern-.05em \textsc{i\kern-.025em b}\kern-.08em
T\kern-.1667em\lower.7ex\hbox{E}\kern-.125emX}}

         \def\UnderWiggleTemp{\the\catcode`\@}
         \catcode`\@=11
         \ifx\UnderWiggle@Loaded\relax
           \catcode`\@=\UnderWiggleTemp
           \endinput \else \let\UnderWiggle@Loaded=\relax \fi
         \newbox\U@BoxA
         \newbox\U@BoxB
         \newdimen\U@DimenA
         \def\U@DoUnderWiggle{
           \offinterlineskip
           \vtop{
             \hbox{\vbox{\copy0}}
             \vskip 1.2pt  
             \vbox to 0.4pt{
               \hbox to\wd0{\hss\char'176\hss}
               \vskip0pt minus 1fil
             }
             \vskip 0.4pt  
           }
         }
         \def\UnderWiggle#1{{%
           \ifmmode
             \mathchoice
               {\setbox0=\hbox{$\displaystyle #1$}\U@DoUnderWiggle}
               {\setbox0=\hbox{$\textstyle #1$}\U@DoUnderWiggle}
               {\setbox0=\hbox{$\scriptstyle #1$}\U@DoUnderWiggle}
               {\setbox0=\hbox{$\scriptscriptstyle #1$}\U@DoUnderWiggle}
           \else
             \setbox0=\hbox{#1}\U@DoUnderWiggle
           \fi
         }}
         \catcode`\@=\UnderWiggleTemp
         \newcommand{\uw}{\UnderWiggle}

\begin{document}

\title{Early Detection of Ovarian Cancer by Wavelet Analysis of Protein Mass Spectra}
\author{Dixon Vimalajeewa, Scott  Alan Bruce, Brani Vidakovic }
\date{}


\maketitle
\section*{Abstract}
Accurate and efficient detection of ovarian cancer at early stages is critical to ensure proper treatments for patients. Among the first-line modalities investigated in studies of early diagnosis are features distilled from protein mass spectra. This method, however, considers only a specific subset of spectral responses and ignores the interplay among protein expression levels, which can also contain diagnostic information. We propose a new modality that automatically searches protein mass spectra for discriminatory features by considering the self-similar nature of the spectra.  Self-similarity is assessed by taking a wavelet decomposition of protein mass spectra and estimating the rate of level-wise decay in the energies of the resulting wavelet coefficients. Level-wise energies are estimated in a robust manner using distance variance, and rates are estimated locally via a rolling window approach. This results in a collection of rates that can be used to characterize the interplay among proteins, which can be indicative of cancer presence. Discriminatory descriptors are then selected from these evolutionary rates and used as classifying features.  The proposed wavelet-based features are used in conjunction with features proposed in the existing literature for early stage diagnosis of ovarian cancer using two datasets published by the American National Cancer Institute.  Including the wavelet-based features from the new modality results in improvements in diagnostic performance for early-stage ovarian cancer detection.  This demonstrates the ability of the proposed modality to characterize new ovarian cancer diagnostic information.

 Keywords: {\it ovarian cancer diagnostics; wavelet spectra; classification; distance variance}

\section{Introduction}
The American Cancer Society (ACS) estimates that 12,810 deaths due to ovarian cancer will occur in 2022, ranking fifth in cancer-related deaths among women in the United States \cite{R8}. According to ACS statistics, a woman's risk of getting ovarian cancer during her lifetime is about 1 in 78, and her risk of death from ovarian cancer is about 1 in 108.  A key factor in reducing mortality risk is early detection, which carries a five-year survival rate of about 89\%. However, most instances of ovarian cancer are more advanced at the time of diagnosis because symptoms in early stages may not be apparent and are similar to other, less serious conditions \cite{R8}.  Also, preventative screening has not been efficacious in prospective randomized controlled trials \cite{R7}.  Only 23\% of ovarian cancer diagnoses are detected early (Stage I) and the five-year survival rates for advanced-stage diagnoses range from 71\% (Stage II) to 20\% (Stage IV). Thus, development of early-detection techniques for ovarian cancer is highly desirable. 

Proteomic pattern analysis is one of the most promising early detection techniques investigated by medical researchers. This involves the collection of serum samples and analysis of serological proteomic patterns, which often reflect pathological changes in an organ or tissue. This is followed by the identification of diagnostic patterns within complex proteomic profiles to characterize normal, benign, and disease states of tissues and body fluids using mass spectrometry. At the proteomic level, such patterns can correspond to specific protein expressions indicative of cancer, helping to differentiate cancer cells from normal cells. For example, cancer antigen 125 (CA125) is a widely used protein biomarker in ovarian cancer diagnosis \cite{R2}. Therefore, exploring expression levels of such proteins can aid in determining pathological stage of ovarian cancer.

Protein mass spectra are generally very spiky and highly complex (see Fig. \ref{fig-0}) due to the the presence of particular ions associated with the serological proteomic profile.  This complexity underscores the need for advanced data mining techniques to explore patterns in mass spectra to uncover pathological changes associated with ovarian cancer. Several studies have explored the diagnostic value of mass spectra using different feature extraction methods relying on genetic algorithms, decision trees, and neural networks. For example, Petricoin et al. \cite{R2}  demonstrated that a genetic algorithm-based feature selection and classification algorithm can detect ovarian cancer with 100\% sensitivity and 95\% specificity. Tang et al. \cite{R17} selected the most informative features from mass spectra using three methods: decision trees, support vector machines, and neural networks, and then investigated differences between the cancer and non-cancer mass spectra. The classification performance reported in this study ranges from 82\% to 99\%. Similarly, a method using the wavelet transform to de-noise protein mass spectra combined with the use of neural networks and discriminant analysis techniques is reported by Vannucci et al.\cite{R1} and achieves similar classification accuracy as the other studies mentioned herein. To summarize these developments, Li et al. \cite{R3} compiled a review of different methods for feature selection and classification used for cancer detection across a variety of cancer types including ovarian cancer. In general, the features created from these studies are based on the magnitude of intensities of protein mass spectra at particular mass-to-charge ratios.  

In all of these studies, extensive pre-processing of mass spectra is conducted comprising baseline correction, peak alignment, normalization, and denoising.  Then, binning or clustering is used to identify ranges of mass-to-charge ratios that have significantly different intensities, and a single mass-to-charge ratio is selected from each bin or cluster.  Finally, various feature selection and classification algorithms are applied to these features to develop the best model for ovarian cancer prediction. There are a variety of approaches to each of these pre-processing steps.  Differences in data pre-processing may contribute to the different mass-to-charge ratios identified as predictive of ovarian cancer across these studies. Accordingly, a modality that requires less pre-processing would be a welcome development and would potentially lead to more consistent and reproducible findings. Additionally, existing studies discard many mass-to-charge ratios across the spectra, which may also contain valuable information.
A modality that considers the full spectra for feature selection without initially clustering or binning is desirable and may lead to further improved diagnostic performance.  

Also, existing studies do not consider differences in associations among mass-to-charge ratios that could also be indicative of ovarian cancer.  One way to characterize such differences is through the the self-similar characteristics of mass spectra, which can capture the interplay between protein expression levels effectively without requiring significant pre-processing. Self-similarity refers to the tendency of signals, in this case protein mass spectra, to exhibit similar properties and behaviors when inspected at different resolutions. These properties have been used often in modeling different natural processes such as landscapes, turbulent flows, and in diagnosing diseases for different areas of diagnostic medicine. For example, Jung et al. \cite{R10} reported that the self-similar properties of Nuclear Magnetic Resonance (NMR) spectra are able to effectively detect plasma cysteine deficiency in the human body. Similarly, Jeon et al. \cite{R4} uses self-similarity in breast cancer images to effectively differentiate images of cancer tissue from healthy tissue. To the best of our knowledge, the value of self-similar properties of protein mass spectra for diagnosing ovarian cancer has not been explored.

This work comprises two important contributions. We first propose a new modality for characterizing self-similar behavior in a manner that is robust to outliers and noise typically present in protein mass spectra. Self-similarity in protein mass spectra is assessed in the wavelet domain.  More specifically, wavelet spectra of the protein mass spectra are computed to characterize regularity of level-wise decay in the resulting scale-specific energies of the wavelet decomposition. The rate of the level-wise decay quantifies the degree of regularity in the mass spectra and can be used as a discriminatory feature.  The energies at different levels are calculated as the level-wise variance of the wavelet coefficients.  However, the traditional variance is known to be sensitive to outliers and noise, which are typically present in protein mass spectra and can negatively impact subsequent diagnostic performance. The primary contribution of the new proposed modality is in the use of the distance variance to estimate level-wise wavelet energies, which has been shown to be robust to outliers and additional noise, and has not been previously used for characterizing wavelet spectra.  In this work, it is shown that the robustness of the new modality leads to improved ovarian cancer diagnostic performance compared to modalities based on traditional variance for characterizing self-similarity.  Additionally, instead of relying on a single slope to characterize the entire mass spectra, we use a rolling window-based approach. The rolling window approach allows for exploring evolutionary slopes and selecting  localized features  that can be used for subsequent detection of ovarian cancer.  

Another contribution of this work is to combine the features derived from the proposed modality with the features defined by existing methods to assess the diagnostic contribution of the new modality in detecting ovarian cancer. Classification performance is assessed using three different classification models, logistic regression (LR), support vector machine (SVM), and k-nearest neighbor (KNN), by reporting their sensitivity, specificity, and correct classification rate (accuracy). By combining the self-similar features from the proposed modality with the magnitude-based features introduced in previous works, we demonstrate improved diagnostic performance compared to the use of the magnitude-based features only. This reflects that, for protein mass spectra, not only the magnitude of expression levels for individual proteins are important, but the interplay among  expression levels of proteins is also critical in detecting ovarian cancer.  

The remainder of the paper is organized as follows: Section \ref{sec-2} gives an overview of the motivating study and datasets used for analysis. The techniques used in this study, fundamentals of wavelet transform,  definition of distance variance, as well as of wavelet spectra, are presented in Section \ref{sec-3}. Sections \ref{sec-4} and \ref{sec-5} contain classification results and discussion respectively, followed by some concluding remarks in Section \ref{sec-6}.

\section{Motivating Study}\label{sec-2}
\begin{figure*}[h]
\centering
 \includegraphics[trim={1cm 7cm 1cm 7cm},clip,width= .7\linewidth]{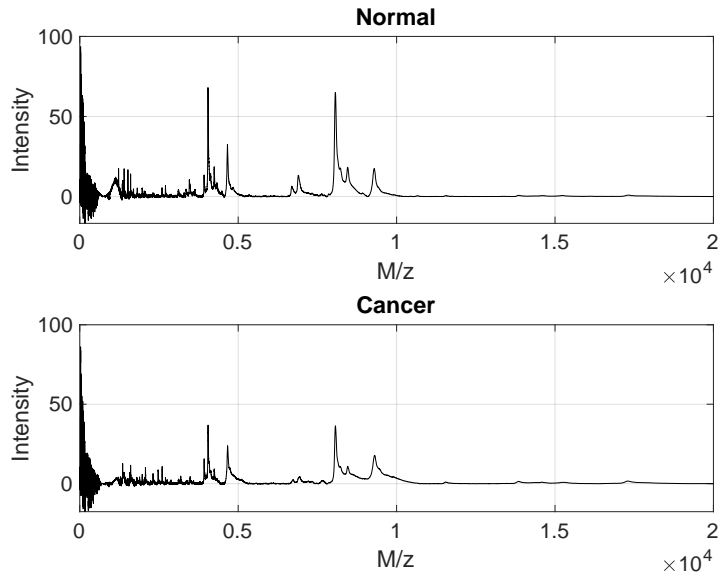}

        \caption{Examples of protein mass spectra from the normal and cancer group.}
\label{fig-0}
\end{figure*}

This study analyzes two ovarian cancer datasets from American National Cancer Institute internet repository \cite{R39}. These datasets consist of protein mass spectra generated from the surface-enhanced laser desorption-ionization time-of-flight (SELDI-TOF) mass spectrometer by using blood samples collected from ovarian cancer patients and healthy controls. The first dataset (designated \emph{Ovarian 4-3-02}) was collected using the weak cation exchange (WCX2) protein chip and PBI SELDI-TOF mass spectrometer. This dataset consists of 216 protein mass spectra generated from 100 cancer, 100 non-cancer, and 16 benign samples. In collecting the second dataset (designated \emph{Ovarian 8-7-02}), the same WCX2 protein chip was used but with an upgraded version of the PBI SELDI-TOF mass spectrometer. Thus, {\it Ovarian} 8-7-02 significantly differs from {\it Ovarian} 4-3-02 in how protein mass spectra were recorded. In this case, the collected protein mass spectra are taken from 162 women diagnosed with cancer and 91 from women in which cancer has not been diagnosed. For both datasets, protein mass spectra consist of the intensities of 15,153 peptides defined by their mass-to-charge ratio (i.e. the ratio of molecular weight to electrical charge) ($M/z$). Figure \ref{fig-0} shows two sample protein mass spectra, one from the healthy control group and the other from the cancer group, selected from {\it Ovarian} 4-3-02. The $x-$axis represents the $M/z$ values, and the $y-$axis displays the spectral intensity.  Interested readers can find more details about these datasets and corresponding studies in \cite{R39,R2}. 

One particular challenge associated with these datasets is the possible presence of non-biological experimental bias in the magnitude of intensities between cancer and control protein mass spectra. Sorace and Zhan \cite{R2a} analyzed the Ovarian 8-7-02 dataset and were able to obtain near-perfect classification accuracy using magnitudes of only two, very low, mass-to-charge ratios (2.7921478 and 245.53704). The inclusion of five more mass-to-charge ratio values resulted in perfect classification accuracy in both training and test sets.  The authors then note that very low mass-to-charge ratios are particularly susceptible to experimental bias due to variation in sample collection, processing, and preservation. For comparative purposes, the magnitudes for very low mass-to-charge ratios (below 500) are not included in subsequent analyses to mitigate the influence of differences between cases and controls unrelated to disease pathology \cite{R2a}. From a modeling perspective, this means that magnitude-based features are particularly sensitive to differences in scalings between protein mass spectra for cases and controls that may be produced by experimental bias.  The proposed wavelet-based features are invariant to such scaling differences and thus provide features that are robust to this particular form of non-biological experimental bias. 

\section{Methods}\label{sec-3}
In this section, we first describe the wavelet transform and wavelet spectra of a signal.  Then, the proposed distance variance-based approach to computing the wavelet spectra is introduced, which represents the methodological novelty of this study, and to the best of our knowledge has not been previously proposed.  Finally, a procedure for feature selection is presented.

\subsection{Wavelet Transform and Wavelet Spectra}
\begin{figure*}[!t]
\centering
    \includegraphics[width= .7\linewidth]{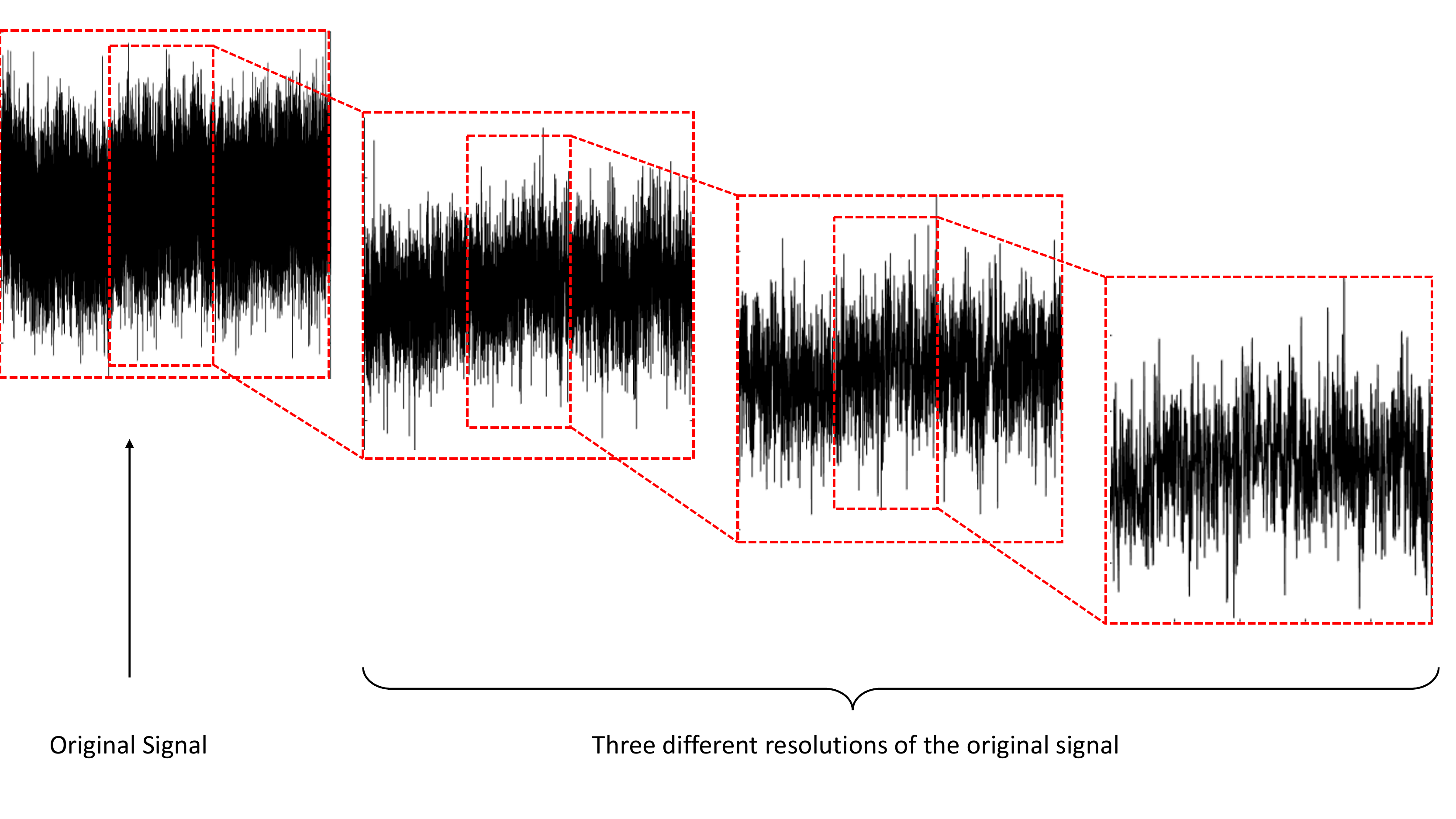}
    \caption{ A sample high-frequency signal at three different resolutions. The signal exhibits similar properties and behaviors (e.g. mean, variance)  when explored the signal at different resolutions. The tendency of exhibiting such similar properties at different resolutions is identified as self-similar nature of the signal.}
    \label{fig-00a}
\end{figure*}

The wavelet transform (WT) is a standard data processing tool commonly used for analyzing complex signals, such as high-frequency time series. The WT decomposes a signal into contributions that are localized both in time and frequency, supporting localized analysis at different resolutions simultaneously. More precisely, wavelet transformed signals present a hierarchy of resolutions, or scales, which in turn help to identify and extract various scale-sensitive properties of signals, such as long memory, fractality, and self-similarity \cite{R21}. Self-similarity is of particular interest in this work and refers to the similarity of the dynamics of a signal at different resolutions.  For example, Fig. \ref{fig-00a} displays a self-similar signal at three different resolutions. Thus, by transforming the signal into the wavelet domain, certain intrinsic properties of the signal, which are difficult to identify in the original domain, are more readily apparent.  As such, the WT is quickly becoming a standard tool for identifying and extracting finer systematic variabilities in complex signals that are difficult to capture. Readers can find more details about the WT and different applications of the WT in Vidakovic \cite{R21} and Morettin \cite{R14}.

\begin{figure*}[!t]
\centering
    \includegraphics[width= .6\linewidth]{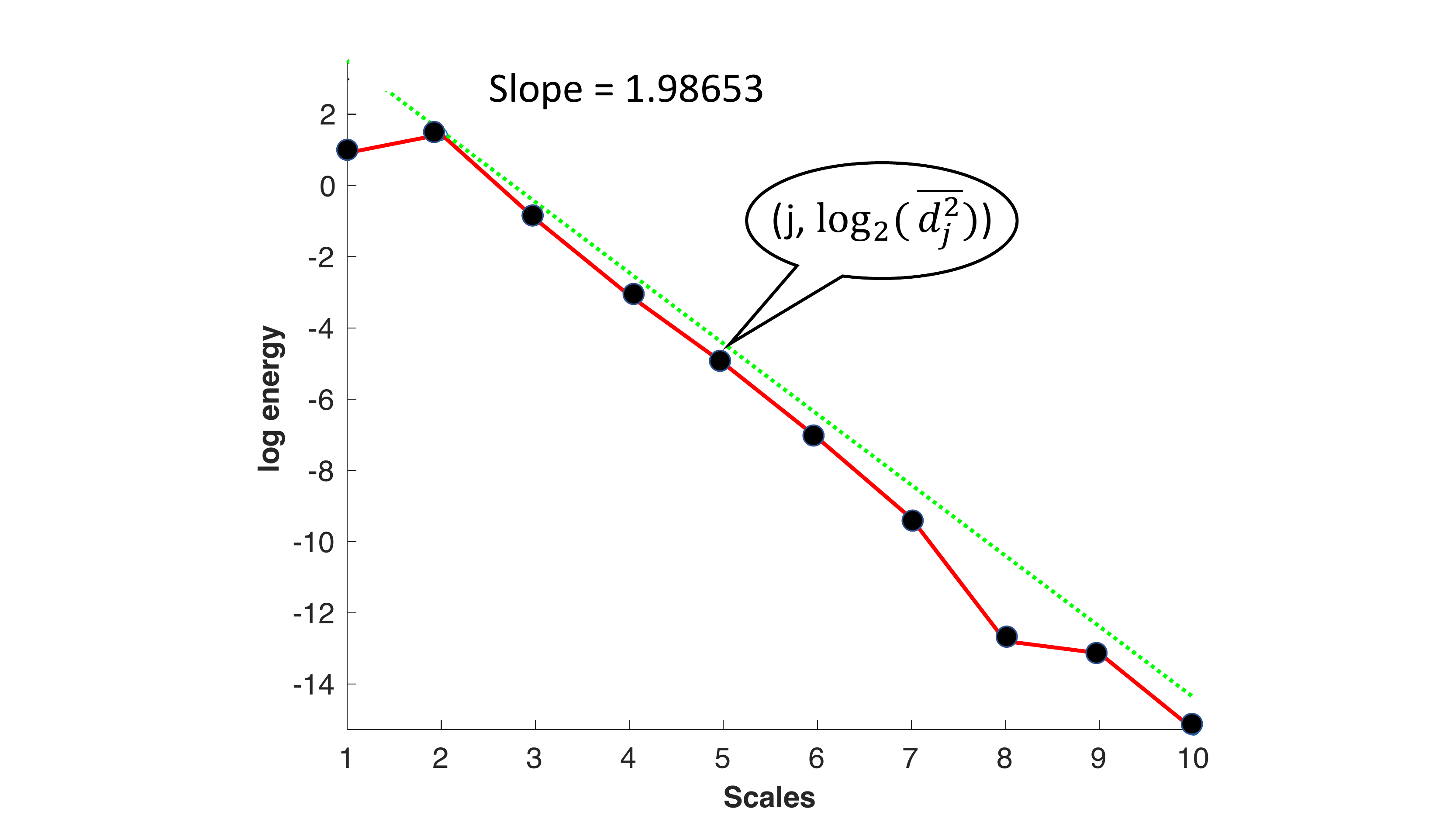}
    \caption{ Wavelet spectra of the signal displayed in Fig. \ref{fig-00a}. Slope of the wavelet spectra is estimated by fitting a straight line (green dashed) on the log energy of the wavelet coefficients (black) within the scale index $j$ ranging from 1 to 10 (red line). The coordinate of the point at the level $j$ is $\log_2(\bar{d_j^2})$, where $\bar{d_j^2}$ is the average-square of the wavelet coefficients at the level $j$ (see Appendix \ref{sec:appa} for more details on wavelet coefficients $d_j$).}
    \label{fig-00}
\end{figure*}

The WT of a given signal results in a set of coefficients (numerical values), which represent the nature of the signal at different resolutions (scales) and locations.  Technical details for computing these coefficients can be found in Appendix \ref{sec:appa}. Using these coefficients to form the wavelet spectra allows for understanding and characterizing the degree of self-similarity of the signal. The wavelet spectra is formed by taking the log average squared of WT detail wavelet coefficients, which are also referred to as log energies, at different scales.  Since the WT details coefficients have zero mean, the wavelet spectra represents the log-variance of wavelet coefficients at different scales. Self-similar signals, such as the signal displayed in Fig. \ref{fig-00a}, exhibit a particular behavior in their wavelet spectra such that the log energies decay linearly as resolution decreases (scale increases) \cite{R10}.
To better illustrate this behavior, Fig. \ref{fig-00} displays the wavelet spectra for the self-similar signal seen in Fig. \ref{fig-00a}.  The rate of energy decay (slope) characterizes the degree of self-similarity in the signal and is estimated by regressing the log energies on the scale indices $j=1,2,\ldots,\log_2(T)-1$ where $T$ is the number of observed time points. For example, the estimated slope of the wavelet spectra shown in Fig. \ref{fig-00} is $-1.98653$.  Larger slopes ($>-2$) indicate a higher degree of self-similarity, and smaller slopes ($<-2$) indicate a lesser degree of self-similarity. Technical details on computing the standard wavelet spectra can be found in Appendix \ref{sec:appb}. Readers can find more details on wavelet spectra in Robert et al. \cite{R34} and Kong et al \cite{R35}. 

Precise estimation of the slope of wavelet spectra is critical for accurately characterizing the degree of self-similarity.  In the context of analyzing protein mass spectra, the high peaks in the protein mass spectra hampers the ability to accurately estimate the slope of the wavelet spectra due to the sensitivity of the traditional mean and variance estimators to outliers. Therefore, we introduce a distance variance-based method in what follows to precisely estimate the log-variance of the energies that make up the wavelet spectra and the corresponding slope in the presence of high peaks typically seen in protein mass spectra.

\subsection{Distance Variance-based Wavelet Spectra}
In computing distance variance-based wavelet spectra, the only difference from the standard variance-based method is that the distance variance is used to estimate the energies in the wavelet spectra instead of the traditional sample variance (average-square of the WT coefficients).  The distance variance of the WT coefficients for each of the scales is computed as follows.

Consider the WT of the signal $Y$ and WT coefficients for the $j$th scale,  $\uw{d}_j = \{d_{1}, d_{2}, \cdots, d_{n}\}$ where $n$ is the number of WT coefficients for the $j$th scale represented in (\ref{eq-12}) in Appendix \ref{sec:appa}. Let $a_{ij} = |d_i -d_j|$ for $i,j = 1, 2, \cdots, n$ denote the absolute difference between elements in $\uw{d}_j$, and $A_{ij}$ denote the double-centered distance

\begin{equation}
        A_{i\ell} = a_{ij} - \frac{1}{n}\sum_{\ell = 1}^n a_{i\ell} - \frac{1}{n}\sum_{m = 1}^n a_{mj} + \frac{1}{n^2}\sum_{m = 1}^n\sum_{\ell= 1}^n a_{\ell m},   \quad  \textrm{for} \quad  i,j = 1,2,\cdots, n \nonumber
\end{equation}

Then, the distance variance of $\uw{d}_j$ is
\begin{equation} \label{eqn-2}
        \mathcal{V}_n^2(\uw{d}_j) = \frac{1}{n^2}\sum_{i = 1}^n \sum_{j = 1}^n A_{ij}^2.
    \end{equation}
In a more general setup, details about the distance covariance can be found in Sz\'ekely et al. \cite{R16} and Sz\'ekely and  Rizzo  \cite{R24}.  The distance variance of a vector is, of course, the distance covariance of the vector with itself.

As many studies reflect, the distance variance and distance covariance are robust counterparts to the traditional sample variance and sample covariance estimators.  These measures are often used to robustify procedures that depend on traditional measures of data dispersion. For example, the study in Matteson et al. \cite{R31} introduces a distance covariance-based independent component analysis (ICA) approach that improves performance over several other computing methods, such as FastICA and kernel density ICA.  Cowley et al. \cite{R32} proposes a dimension reduction technique based on the distance covariance. This study showed that the distance covariance approach can achieve better performance compared to existing dimension reduction methods, such as principal component analysis and canonical component analysis.  A more recent study by Emmanuel et al. \cite{R33} proposes the distance covariance based minimum dependence estimator (MDep) and demonstrates that this estimator improves estimation accuracy compared to ordinary least squares in the context of instrumental variables regression. Similarly, the studies in Sz\'ekely and Rizzo \cite{R25}, \cite{R26} and Li and Liping \cite{R27} introduce improved robust techniques for measuring dependencies and screening features in complex data that are based on the notion of distance covariance.

\begin{figure}[ht]
\centering
\begin{subfigure}{.52\linewidth}
  \centering
  \includegraphics[width=1\textwidth]{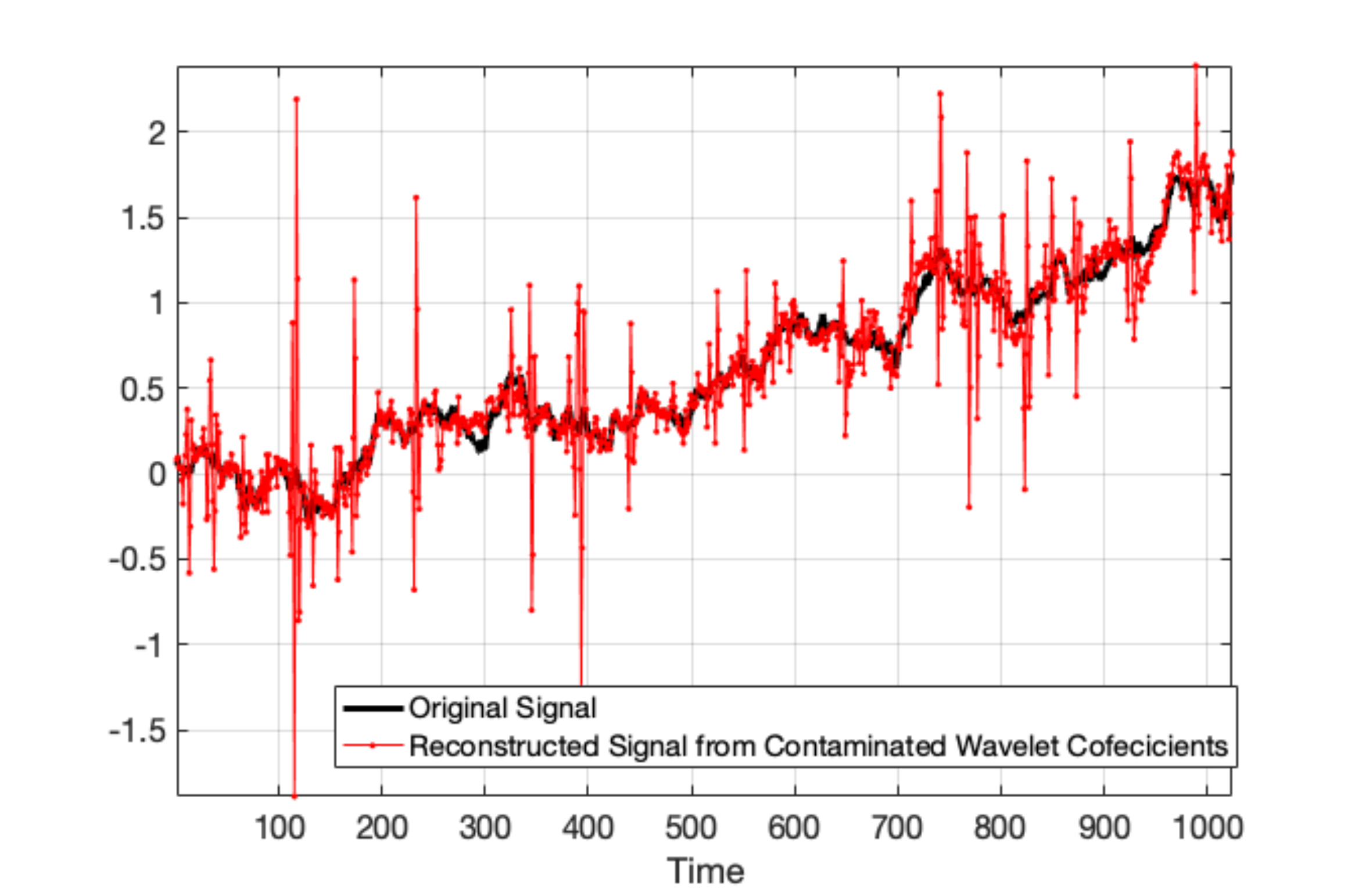}
  \caption{}
  \label{fig-41}
\end{subfigure}
\begin{subfigure}{.47\linewidth}
  \centering
  \includegraphics[width= 1\textwidth]{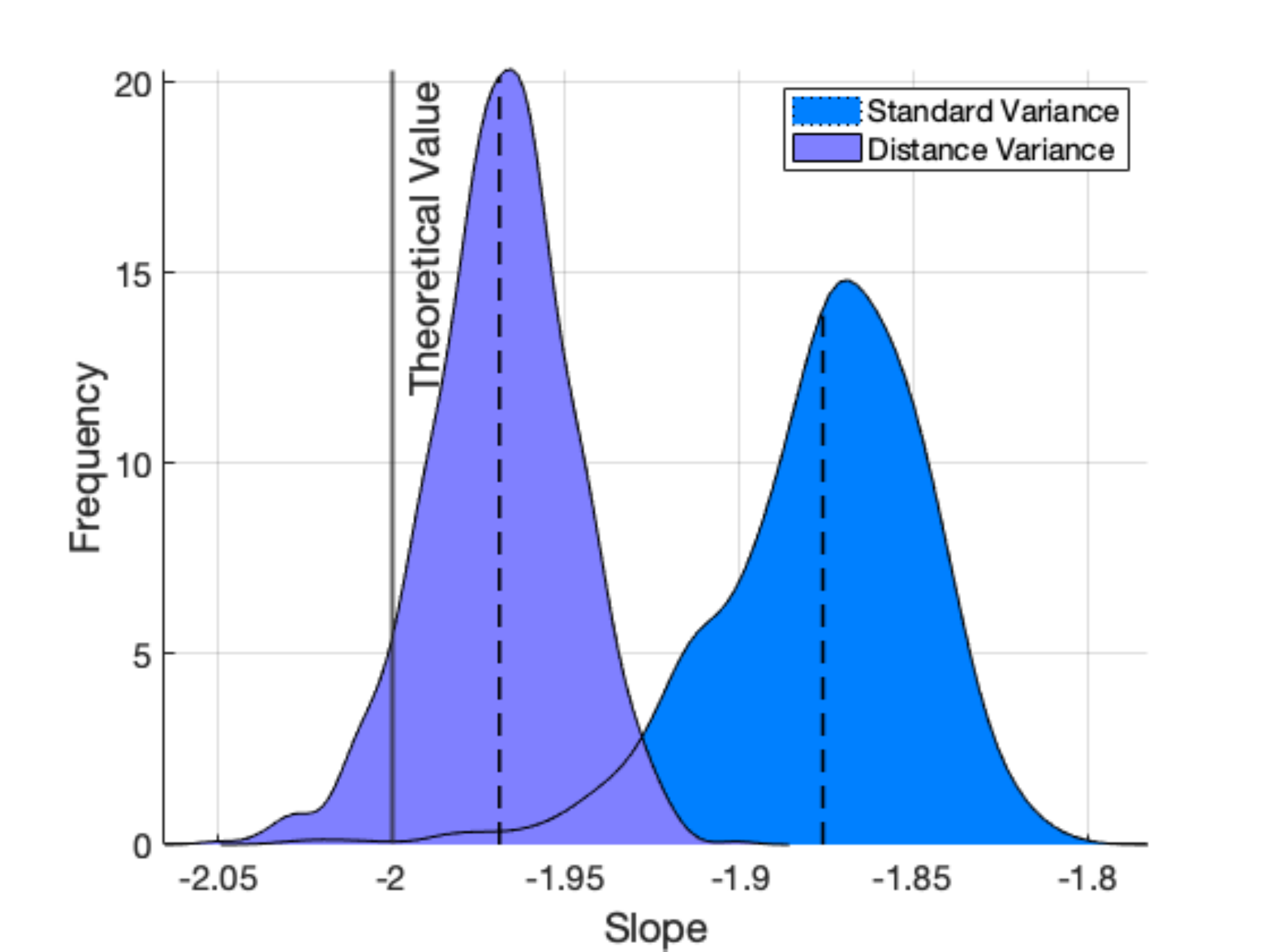}
  \caption{}
  \label{fig-42}
\end{subfigure}

\caption{Effectiveness of estimating slope in wavelet spectra using the standard and the distance variance-based methods for 1,000 realizations of a contaminated Brownian motion process. For each realization, a signal of length 1024 is generated from a standard Brownian motion, and randomly selected WT coefficients are then contaminated by adding standard Gaussian noise.  An example can be found in (a). Slopes of the wavelet spectra are then computed using the traditional variance and distance variance approaches. The dashed lines represent the averages of the estimated slopes using each approach, and the solid line  at $-2$ represents the theoretical slope for a standard Brownian motion process (b).}
\label{fig-04}

\end{figure}

To better illustrate the advantages of using the robust distance variance in this work, the following numerical example compares the traditional sample variance and distance variance estimators of the slope of the wavelet spectra for a Brownian motion signal contaminated with outliers. First, a sample signal of length $T=1024$ is generated from a standard Brownian motion process, and then WTs are taken using the Daubechies-6 wavelet with a decomposition levels 9. Second,  at each resolution level, a few wavelet coefficients are randomly selected and contaminated by adding standard Gaussian noise before computing the wavelet spectra.  The sample variance and the distance variance are then used to estimate the slope.  Finally, the distribution of estimated slopes was obtained from independent 1,000 realizations of the contaminated Brownian motion process described above. The theoretical slope for a standard Brownian motion process is known to be $-2$.  As can be seen in Fig. \ref{fig-04}, the slope estimated using the distance variance approach is much closer to the true value compared to the estimates using the traditional variance. That is, the distance variance-based method is significantly less biased in estimating the slope in the presence of outliers, which demonstrates the general robustness properties of this estimator.

A potential drawback for the use of distance variance is its high computational cost for large datasets \cite{R27}. The computational cost of computing the sample distance variance as it is defined in (\ref{eqn-2}) is $O(n^2)$. If calculated as in (\ref{eqn-2}), the quadratic computational complexity greatly limits the applicability of the distance covariance on large datasets. To alleviate this, fast algorithms have been proposed. For example, the study by Huo and Sz\'ekely \cite{R23}  proposed an algorithm with computational complexity $O(n \log n)$. Chaudhuri and Hu \cite{R15} also proposed an algorithm with the same computational complexity $O(n \log n)$ but significantly faster. For example, this algorithm takes around 4 seconds on a standard personal computer with Intel Xeon Gold CPU 2.40 GHz processor and 16 GB memory to calculate distance variance using one million data points. In this study, we use the fast distance covariance algorithm proposed by Chaudhuri and Hu \cite{R15}.  


\subsection{Feature Extraction}\label{sec-33}
In addition to the more sophisticated binning and clustering approaches to feature extraction described in Section 1, a simpler direct feature extraction method has been used in many studies to analyze the aforementioned ovarian cancer datasets \cite{R3}. This method uses the whole mass spectra to select discriminatory features with large differences in mean values between cases and controls characterized by Fisher's criterion.  

\begin{equation}\label{eqn-3}
            F = \frac{(\mu_{case} - \mu_{control})^2}{\sigma_{case}^2 + \sigma_{control}^2},
        \end{equation}
where $\mu_{case}$ and $\mu_{control}$ are the arithmetic mean for the intensities at each mass-to-charge ratio for the case and control groups, respectively,  and $\sigma_{case}^2$ and $\sigma_{control}^2$ are the variances of the corresponding intensities over the mass-to-charge ratios in the mass spectra for the case and control groups.  This measure is computed for all 15,153 mass-to-charge ratios in the mass spectra. Then the mass-to-charge ratios with the highest $F$ values are selected for subsequent classification purposes. Prior studies indicate that using 10 mass-to-charge ratios with the largest $F$ values can achieve good classification performance \cite{R3}. 

This study, however, uses an evolutionary spectra-based approach for deriving discriminatory features from the mass spectra. The evolutionary spectra-based method uses a moving data window to select features. That is, discriminatory features are selected from a data window while moving the data window along the mass spectra with a specific step size. Thus, this approach enables capturing more localized features in the mass spectra while still accounting for intensities of mass-to-charge-ratios. Spectral slopes of the wavelet spectra of those data windows are extracted as discriminatory descriptors. Depending on the step size used to move the data window along the mass spectra, those descriptors could be extracted from overlapping (or non-overlapping) data windows. This study uses overlapping data windows, aiming to extract as many discriminatory features as possible from the mass spectra. 

\begin{figure*}[!t]
\centering
 \includegraphics[width= 1\linewidth]{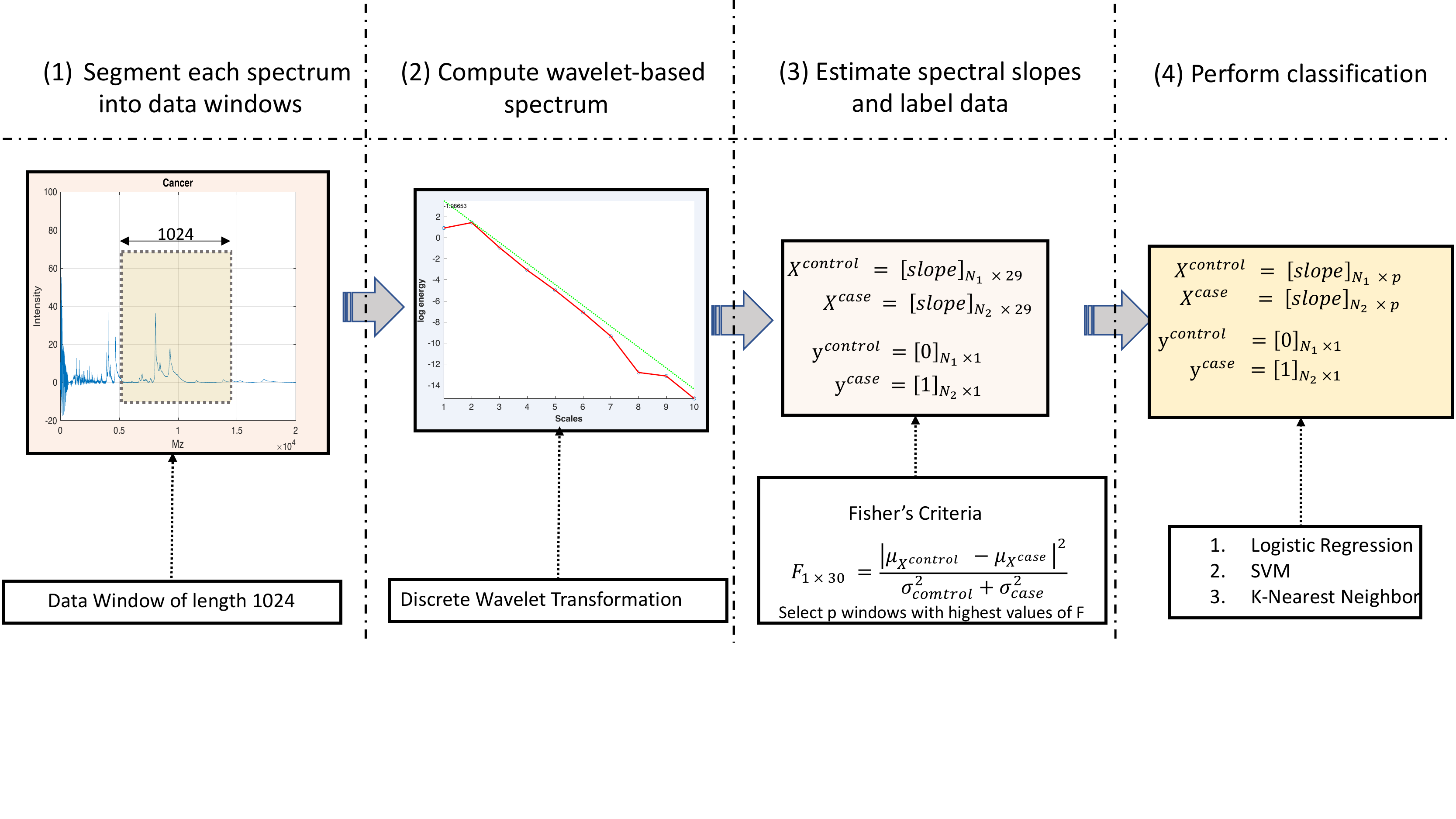}
 \vspace{-8.0em}
        \caption{An overview of the ovarian mass spectra classification process using the wavelet spectra with a dataset which consists of $N_1$ and $N_2$ number of mass spectra in the case and control groups, respectively. Each spectrum in the case and control group is divided into a set of data windows of length 1024 (29 data windows in all) (1). Then, the estimated slope of the wavelet spectra for each data window is computed (2). Feature matrices ($X$) for the case and control groups are formed using the estimated slopes for the $p$ windows with the largest Fisher's criteria (3); finally, slopes are used  to fit classification models (4).}
\label{fig-01}
\end{figure*}

Fisher's criterion (\ref{eqn-3}) is then used to select a set of the most informative data windows in the mass spectra that well discriminate the cases and controls. To compute the $F$ value for each data window, the mean and variance ($\mu$ and $\sigma^2$) of slopes of each data window corresponding to the case and control groups are used. Then, a set of data windows that correspond to the highest $F$ values are selected as the most informative discriminatory features.  To better understand the rolling window approach, Fig. \ref{fig-01} provides an illustration of the overall feature selection and classification process used for ovarian cancer prediction in this work.
\section{Ovarian Cancer Spectra Classification}\label{sec-4}
This section evaluates the performance of the proposed modality in detecting ovarian cancer. First, we compare classification accuracy using the standard variance- and distance variance-based wavelet features. Second, we investigate the contribution of this modality in detecting ovarian cancer by integrating wavelet-based features derived from the proposed modality with the magnitude-based features derived from the the direct method.

\subsection{Data Preparation} \label{sec-41}
The following data preparation process (Fig. \ref{fig-01}) was performed on {\it Ovarian} 4-3-02 and 8-7-02 to produce both the standard variance- and distance variance-based wavelet features.

\begin{itemize}
    \item [(1)] A data window of length 1024 is selected and the discrete WT is computed for each of the protein mass spectra within this window with the Daubechies-6 wavelet with 6 decomposition levels.
    \item [(2)] The wavelet-based spectrum is computed from the WT coefficients obtained in Step (1), and the slopes of the wavelet spectra are estimated.

    \item [(3)] Wavelet-based features are selected in the following manner
    \begin{itemize}
        \item [(a)] Steps (1) and (2) are repeated while shifting the data window by 500 data points along the protein mass spectra. To cover nearly all of the 15,153 mass-to-charge ratios, $\lfloor (15153-1024)/500 + 1 \rfloor = 29$ overlapping windows are required and cover 15,024 mass-to-charge ratios (i.e.,  29 spectral slopes). The largest 129 mass-to-charge ratios are not included in the analysis. Two feature matrices are then formed using the estimated slopes, and they are denoted as $ X_{N_1 \times 29}^{case}$ and $X_{N_2 \times 29}^{control}$, where $N_1$ and $N_2$ is the number of spectra in the case and control groups, respectively.

        \item [(b)] To select most significant classifying features, Fisher's criterion ($F$) given in Equation (\ref{eqn-3}) is computed, and the $p$ features with the largest $F$ values from the feature matrices are then selected for classification purposes.  This results in smaller feature matrices $ X_{N_1 \times p}^{case}$ and $X_{N_2 \times p}^{control}$ where $p \le 29$.

        \item [(c)] Response matrices  are created by assigning labels 1 and 0 for cases and controls respectively ($y_{N_1 \times 1}^{case}$ and $y_{N_2 \times 1}^{control}$).
       \end{itemize}
    \end{itemize}

\subsection{Model Fitting}\label{sec-42}
For model fitting, 67\% of the rows from each of the feature matrices, $X^{case}_{N_1 \times p}$ and $X^{control}_{N_2 \times p}$, are randomly selected and used for training; the samples in the remaining rows are used for testing. Although {\it Ovarian} 4-3-02 has an equal number of cases and controls (i.e., $N_1 =100$ cases and $N_2=100$ controls), {\it Ovarian} 8-7-02 has $N_1 =162$ cases and $N_2 =91$ controls. To avoid bias due to the imbalance in group sizes for {\it Ovarian} 8-7-02, 91 of the 162 cases are randomly selected to match the number of controls prior to randomly determining the training and test sets. We perform our experiment with three commonly used classification algorithms, logistic regression (LR), support vector machine (SVM), and k-nearest neighbors (KNN); their performance was assessed by computing sensitivity, specificity, and overall correct classification rate (accuracy). This model training and testing process is repeated 10,000 times, and reported performance measures are averaged over the number of repetitions.

\begin{figure*}[!t]
\centering
 \includegraphics[width= .7\linewidth]{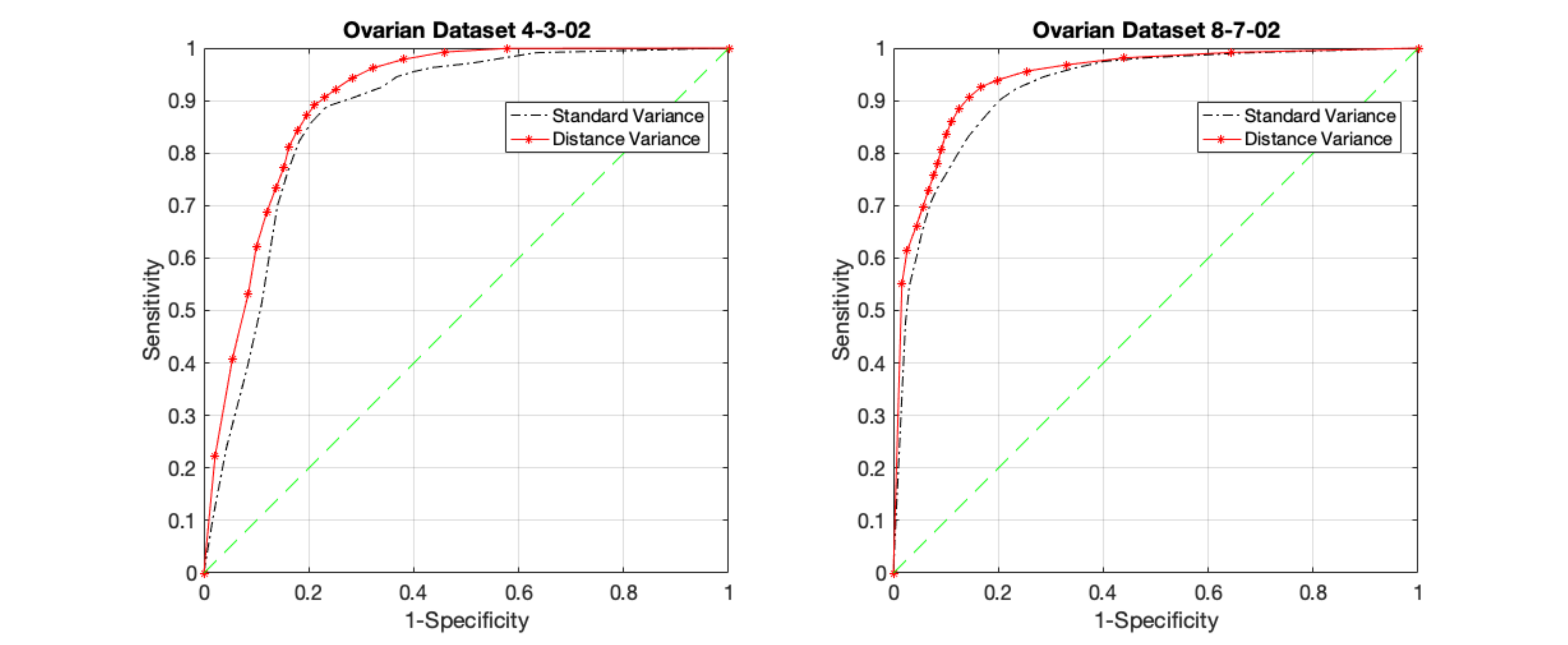}
        \caption{Receiver operating characteristic (ROC) curves computed using the standard variance and distance covariance-based spectral feature  matrices computing methods for {\it Ovarian} 4-3-02 and {\it Ovarian} 8-7-02.}
\label{fig-000}
\end{figure*}

When performing LR, it is essential to choose the best threshold on model predicted probabilities that separates cases and controls. We use the threshold that maximizes the Youden Index, $\gamma = (1/\sqrt{2})(\mbox{\it sensitivity} + \mbox{\it specificity} -1).$  This index is defined as the point on the receiver operating characteristic (ROC) curve which is most distant from the diagonal \cite{R18}. For instance, Fig. \ref{fig-000} shows the ROC curves computed for {\it Ovarian 4-3-02} and {\it Ovarian 8-7-02} for a single repetition. Under the standard and distance variance methods, the best threshold values were 0.53 and 0.42 for Ovarian 4-3-02, and 0.38 and 0.42 for {\it Ovarian} 8-7-02 respectively. The sensitivity, specificity, and accuracy were reported corresponding to these thresholds.

\subsection{Classification Performance of the Proposed Wavelet-based Features}
\begin{figure}[!t]
\centering
\begin{subfigure}{1\textwidth}
  \centering
  \includegraphics[width=.75\textwidth]{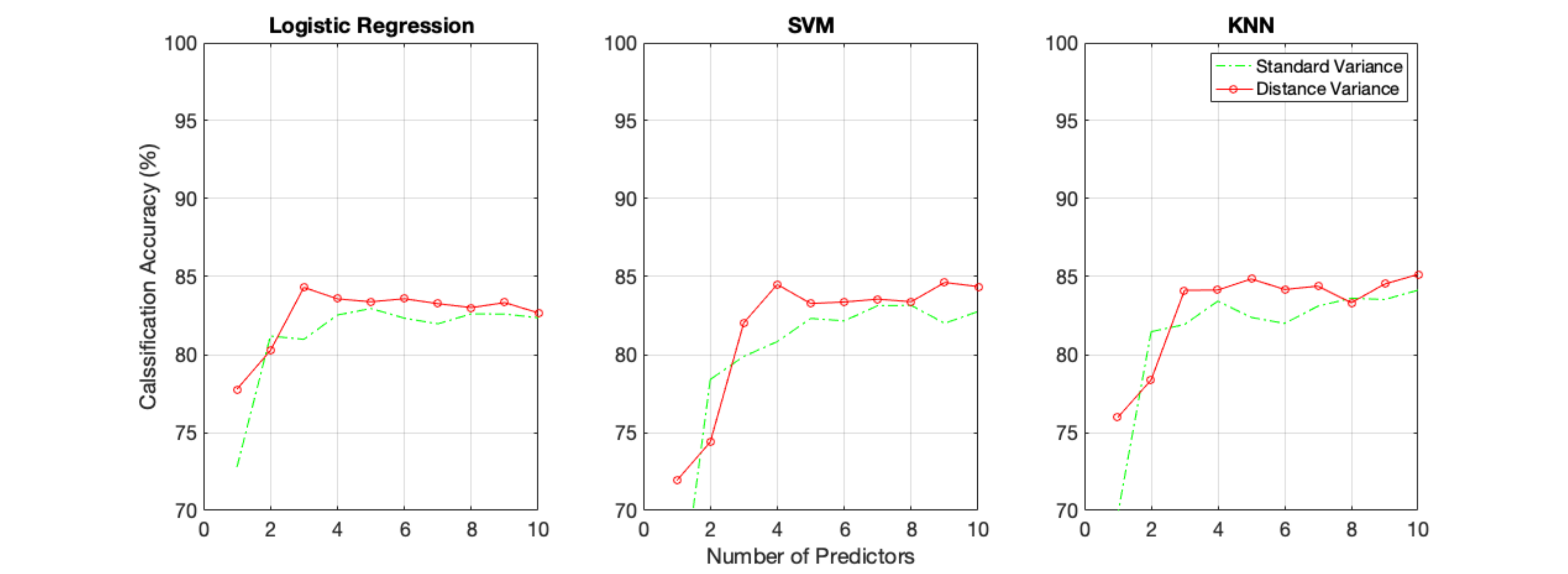}
  \caption{{\it Ovarian} 4-3-02}
  \label{fig-21}
\end{subfigure}
\begin{subfigure}{1\textwidth}
  \centering
  \includegraphics[width= .75\linewidth]{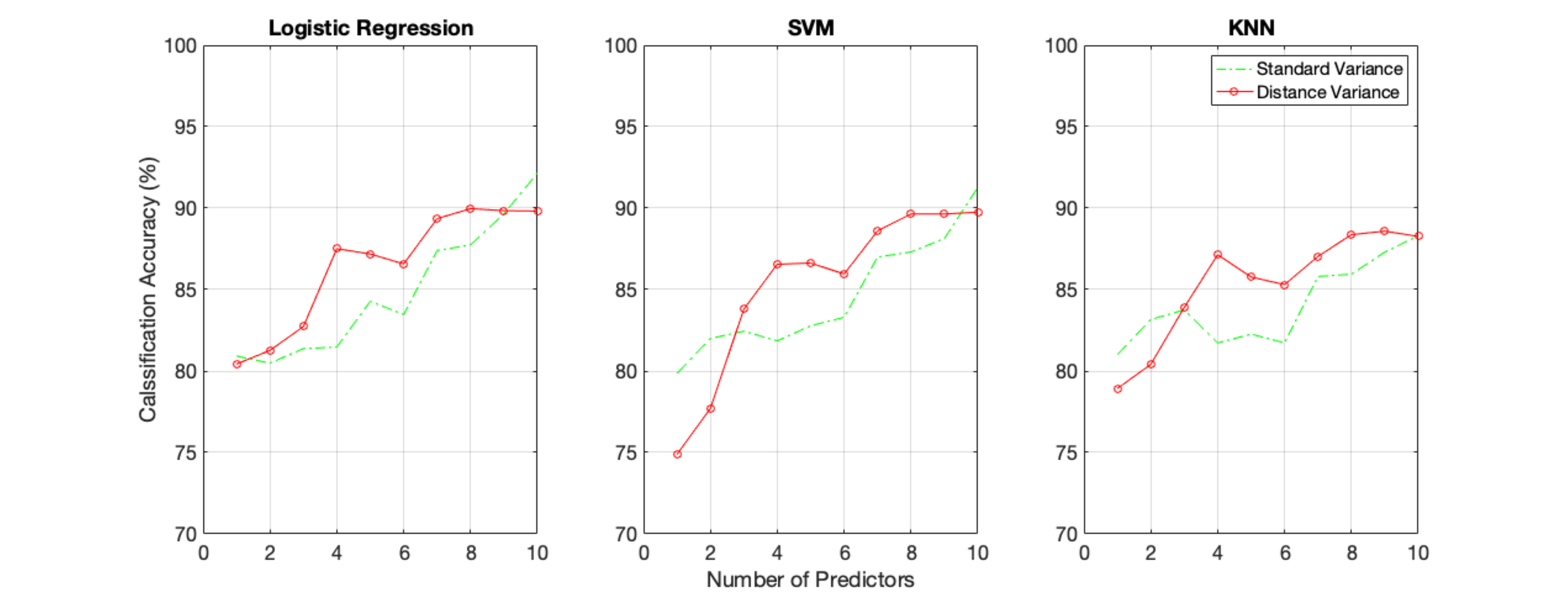}
  \caption{{\it Ovarian} 8-7-02}
  \label{fig-31}
\end{subfigure}
\caption{Change in test set classification accuracy over 10,000 repetitions for different numbers of predictors (selected features) used to fit classification models.}
\label{fig-2}
\end{figure}

We investigated classification performance using the five most significant features corresponding to five particular windows in the evolutionary spectra (Table \ref{tab-1}).  The use of only five features was determined by investigating classification performance for different numbers of features (Fig. \ref{fig-2}). For both the standard and distance variance approaches, classification accuracy on the test sets generally increases as the number of predictors increases, but there is not much improvement in classification accuracy using more than five predictors. Thus, in a spirit of Occam's razor compromising between accuracy and simplicity of the model, five most significant features with respect to Fisher's criterion are used.
Overall, the distance variance-based method shows better classification performance compared to the standard variance method (Table \ref{tab-1}).  Sensitivity analysis on the number of selected features (not shown) indicates that the improved performance offered by the distance variance-based method still holds for different numbers of selected features.

\begin{table*}[!t]
	\centering
	
	\begin{tabular}{llcccccc}
		\hline
		\multicolumn{2}{l}{} & \multicolumn{6}{c}{Method used to compute wavelet spectra}  \\
		\cline{3-8}
		\multicolumn{2}{l}{} & \multicolumn{3}{c}{Standard variance} & \multicolumn{3}{c}{Distance variance} \\
		\cline{3-8}
	    & Model & Sensitivity & Specificity & Accuracy & Sensitivity & Specificity & Accuracy \\
		\hline
	    \multirow{3}{*}{{\it Ovarian} 4-3-02} & LR   & 83.98 & 77.55 & 80.77 & {\bf 88.27} & {\bf 80.00} &                             {\bf 83.10} \\
	                               & SVM  & 80.67 & 79.05 & 79.86 & {\bf 83.88} & {\bf 80.15} & {\bf 83.66} \\
	                               & KNN  & 84.03 & {\bf 79.47} & 81.75 & {\bf 89.35} & 78.72 & {\bf 82.80} \\
		\hline
	    \multirow{3}{*}{{\it Ovarian} 8-7-02} & LR   & 80.78 & 86.29 & 86.55 & {\bf 85.33} & {\bf 88.34} &                             {\bf 88.75} \\
	                               & SVM  & 80.50 & {\bf 86.01} & 86.47 & {\bf 87.13} & 85.23 & {\bf 88.20} \\
	                               & KNN  & 72.12 & 91.08 & 84.72 & {\bf 78.81} & {\bf 91.97} & {\bf 86.95} \\
		\hline
	\end{tabular}
	\vspace{.1cm}
	\caption{Test set classification performance with the standard and distance variance-based methods over 10,000 repetitions. Bold values represent the larger value between the two methods.}
	\label{tab-1}
\end{table*}

\subsection{Classification Performance Using Both Traditional and Wavelet-based Features}
First, classification performance using features derived from the direct method (discussed in Section \ref{sec-33}) is assessed using {\it Ovarian} 4-3-02 and {\it Ovarian} 8-7-02 (Table \ref{tab-5}). Irrespective of the classification algorithm, features from the direct method achieve better classification performance than using only the wavelet-based features from the proposed modality. To assess the contribution of the proposed modality in detecting ovarian cancer, we performed classification by combining the proposed wavelet-based features with the features selected from the direct feature extraction method. As we see in Table \ref{tab-5}, using these features together improves test set classification performance compared to using each set of features individually. Also, the classifiers using features derived from the direct method in conjunction with distance variance-based wavelet features offers better performance compared to classifiers using both direct method features and standard variance-based wavelet features. Thus, the proposed modality contributes to improving ovarian cancer detection performance over existing methods by incorporating new diagnostic information not present in existing features.

\begin{table*}[!t]
	\centering
	
	\begin{tabular}{lllccccccc}
		\hline
		Feature  & CL         & Model & \multicolumn{3}{c}{{\it Ovarian} 4-3-02} & \multicolumn{3}{c}{{\it Ovarian} 8-7-02} \\
		Selection & Method &           &                                                                &\\
		\hline
		& &  & Sensitivity & Specificity & Accuracy &  Sensitivity & Specificity & Accuracy \\
			\cline{4-9}
		& \multirow{3}{*}{Direct} & LR   & 96.44 & 97.90 & 97.17 & 96.83 & 97.95 & 97.50 \\
	                            && SVM  & 96.35 & 97.52 & 96.93 & 96.37 & 97.90 & 97.41 \\
	                            && KNN  & 84.50 & 93.74 & 89.12 & 95.75 & 94.50 & 97.09\\
		\hline \hline
		\multirow{2}{*}{Standard} & Direct & LR  & 96.95 & 97.85 & 97.40 &  99.06 & 99.33  & 99.19  \\
		                          & +      & SVM & {\bf 97.20} & {\bf 98.05} & {\bf 97.63} &  98.78 & 99.39 & 99.08  \\
		                Variance  & Slope  & KNN & 86.45 & 94.10 & 90.27 & 95.28 & 98.61 & 96.94  \\
		 \hline
		\multirow{2}{*}{Distance } & Direct & LR  & {\bf 97.20} & {\bf97.95} & {\bf97.57} &  {\bf99.39} & {\bf99.39} &  {\bf99.39}  \\
		                          & +       & SVM & 97.05 &  97.90 &  97.48 & {\bf 99.44} & {\bf 99.50} & {\bf 99.47}  \\
		                 Variance & Slope   & KNN & {\bf 86.55} & {\bf 94.50} & {\bf 90.52} & {\bf 95.78} & {\bf 98.72}  & {\bf 97.25}  \\

		\hline
	\end{tabular}
	\vspace{.1cm}
	\caption{Test set classification performance of the direct feature extraction method with ten features. A joint approach is achieved by combining the direct and spectral slope-based feature extraction methods. Bold values represent the larger value among the three approaches (direct, direct and standard variance slopes, direct and distance variance slopes).}
	\label{tab-5}
\end{table*}

\section{Discussion}\label{sec-5}

This work introduces a new modality for ovarian cancer detection and shows that it can be used in combination with existing features to improve accuracy in detecting ovarian cancer. This performance improvement is due to the joint use of independently generated feature sets. That is, the proposed modality derives features in the wavelet domain, while the direct method derives features in the original data domain. Existing studies support the combined use of features from multiple domains to better characterize differences among groups leading to improved classification accuracy compared to individual feature extraction methods \cite{R40, R41, R42}. 

\begin{figure*}[!t]
\centering
 \includegraphics[width= 1\linewidth]{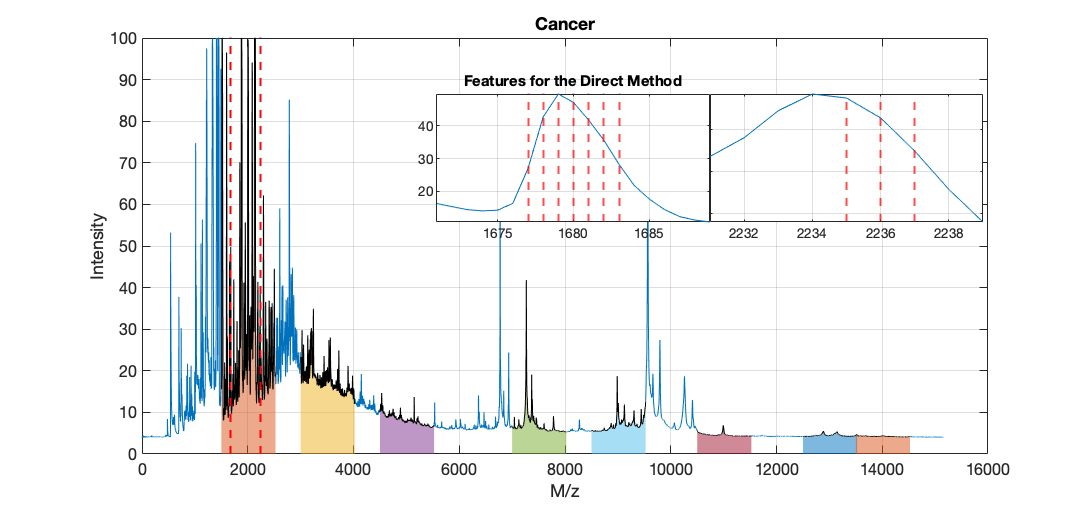}
        \caption{Protein mass spectra for an individual sample from the ovarian cancer group in {\it Ovarian} 8-3-02.  Highlighted ranges of mass-to-charge ratios reflect the ranges for which wavelet-based features are determined to be significant in classifying ovarian cancer using the distance variance approach.  Red dashed lines are 10 particular mass-to-charge ratios identified as significant in classying ovarian cancer using the direct method (i.e. large Fisher's criterion for difference in mean intensities).}
    \vspace{-1.5em}
\label{fig-05}
\end{figure*}

As reported in previous studies, there are interactions among different proteins in the presence of ovarian cancer. That is, the expression of some proteins are up-regulated, while some proteins are down-regulated in the presence of ovarian cancer. For instance, the study by Li et al. \cite{R36} reported lower intensities for a mass-to-charge ratio of 3,759, while the mass-to-charge ratios 4,659 and 9,318 exhibited higher intensities in the presence of cancer. This suggests the presence of interactions between different protein expression levels in the protein mass spectra. Although capturing such interactions is critical for the detection of ovarian cancer, existing features focus only on peaks in the protein mass spectra and differences in peaks between case and control groups, without directly considering interactions.  Whereas, the proposed modality captures such interactions in protein mass spectra across mass-to-charge ratios in the wavelet domain and seeks to mitigate the influence of peaks by introducing a robust distance variance estimator of the scale-specific energies that make up the wavelet spectra. Thus, the existing features and the features defined through the proposed modality capture two different types of characteristics of the protein mass spectra.  To better show the differences between the information captured by the two modalities, see Fig. \ref{fig-05}, which displays the windows of mass-to-charge ratios (colored regions) for which the wavelet-based features selected by the proposed method differ from those of the direct method (red dashed lines). The features derived using the direct method are concentrated in a narrow range of mass-to-charge ratios (1,675 to 1,685 and 2,234 to 2,238), while many other ranges are discovered using wavelet-based features.  This indicates that different characteristics of the protein mass spectra are embedded within each of the two modalities. This is why combining these sets of features improves classification beyond using each individually.

The proposed modality also automatically considers the entire spectra and requires minimal pre-processing. Whereas, the existing features rely heavily on pre-processing of the protein mass spectra (e.g. baseline correction, peak alignment, normalization, denoising, clustering), and there is currently no widely-accepted standardized set of procedures for pre-processing mass spectra. As a result, this hampers the generalizability and reproducibility of important features due to differences in pre-processing procedures. The proposed modality overcomes such issues, as it can achieve comparable classification performance to the existing methods with only minimal pre-processing. This is one of the key advantages of the proposed modality compared to the ovarian cancer detection techniques that have been previously used. 

Another advantage of the proposed modality is that it uncovers new sets of mass-to-charge ratios not previously identified as important, in addition to identifying significant mass-to-charge ratios previously discovered. 
As can be seen in Fig. \ref{fig-05}, the mass-to-charge ratio ranges 7,000-8024, 12,500-13,524, and 13,500-14,524 are not previously found in the literature. This means that no specific biomarkers (as in significant mass-to-charge ratio values) have been identified in these mass-to-charge ratio ranges in previous studies. This is not surprising, since the proposed modality generates features based on a wavelet domain representation, which hasn't been used previously for analyzing protein mass spectra to detect ovarian cancer.  These new ranges of mass-to-charge ratios represent preliminary findings that should be further investigated in future works.

Other mass-to-charge ratio regions contains at least one key potential biomarker which have been proposed in previous studies. For example, Petricoin et al. \cite{R2} reported that based on {\it Ovarian} 4-3-02 and 8-3-02, the optimum discriminatory patterns in the ovarian mass spectra are at the mass-to-charge ratio ($M/z$) values 534, 989, 2,111, 2,251, and 2,465. A sensitivity of 100\% and a specificity of 95\% were reported with these features. The last three mass-to-charge ratio values 2,111, 2,251, and 2,465 belong to the mass-to-charge ratio region 1,500-2,524 detected by our modality. Similarly, three potential peaks, 3759, 4,659, and 9,318, were reported to be the most important mass-to-charge ratio values in detecting ovarian cancer in Li et al. \cite{R36} These values include in the mass-to-charge ratios 3,000-4,024, 4,500-5,525, and 8,500-9,524, respectively. Zang et al. \cite{R37} detected serum protein markers at  peak mass-to-charge ratio value 4,475, 6,195, 6,311, 6,366, and 11,498. In this case, our model can only detect 4,475 (4,500-5,524) and  11,498 (10,500-10,524). The study by Fung et al. \cite{R38} identified a significant  biomarker, inter-alpha tryp-sin inhibitor heavy chain 4 (ITIH4), which peaks at a mass-to-charge ratio of 3,272. This biomarker belongs to the mass-to-charge ratio region 3,000-4,024 detected as one of the most discriminatory features by the proposed modality. Overall, the proposed modality can automatically capture several previously identified and also new mass-to-charge ratios that can serve as biomarkers that are crucial for the detection of ovarian cancer.

It is important to note that the study by Sorace and Zhan \cite{R2a} emphasized that the selection of features below a mass-to-charge ratio of 500 may be the result of significant non-biological experimental bias. As a result, the use of features within this range should be carefully assessed when using for ovarian cancer detection generally.  As can be seen in Fig. \ref{fig-05}, the proposed modality does not detect any significant mass-to-charge ratios within this range for the detection of ovarian cancer. Thus, this demonstrates another advantage offered by the proposed modality, which is invariant to scale differences between groups that may be due to non-biological experimental bias.

\section{Conclusions}\label{sec-6}

This paper proposes a new modality based on self-similarity in the protein mass spectra and demonstrates its performance in ovarian cancer diagnostics. Self-similar property in data signals is assessed in the wavelet domain by computing wavelet-based spectra of protein mass spectra. In the abstract we referred to ``spectra of the spectra,'' that is protein mass spectra is analyzed by calculating its wavelet spectra. We hope that there are no confusions in the dual use of word {\it spectra}, and that the connotation is clear from the context.

The new modality introduced in this work uses a distance variance-based calculation of the wavelet spectra and investigates its performance of extracting discriminatory features. This is followed by exploring the contribution of the extracted features towards improving ovarian cancer detection by using these features together with existing features  discovered previously.

Regardless of the classification algorithm used, the classifying features derived using the distance variance-based method have better classification performance compared to the method using the standard sample variance. For both datasets, the classification accuracy varies within the range 79 - 87 \%  and 83 - 89 \% for the proposed modality based on the standard and distance variance-based methods, respectively. The joint use of features selected through the proposed modality with the features of existing methods contributes to overall improved detection (i.e., classification accuracy $90 - 99\%$), compared to the performance of individual feature extraction methods. The proposed modality captures, with minimal pre-processing, new significant mass-to-charge ratio ranges along with mass-to-charge ratio ranges covering the majority of biomarkers discovered previously.   

However, there are some limitations of the proposed modality that could be addressed in future research. For instance, the distance variance approach minimizes the influence of peaks in the protein mass spectra but does not eliminate their influence entirely. Mixture models could be developed to simultaneously characterize peaks and dependencies (unrelated to these peaks) across the spectra in order to produce features based on differences in peaks, as well as wavelet-based features capturing dependencies \cite{R43}. It is also important to account for machine effects and measurement noise in modeling efforts, which can negatively impact classification performance. This would require exploring models with careful fine-tuning to mitigate influence of such nuisance information. For instance, the study by McNamara et al. \cite{R44} reports possible machine learning methods to overcome these issues. 

In the spirit of reproducible research, the software used in this paper is posted on \url{https://web.stat.tamu.edu/~brani/wavelet/}.

\appendix

\section{Wavelet Transform}\label{sec:appa}
The discrete wavelet transform (DWT), a popular version of wavelet transforms (WTs), has emerged as an important tool for analyzing complex data signals in application domains where discrete data are analyzed. Simply, DWTs are linear transforms that can be represented by orthogonal matrices. Suppose a data signal ($Y$) of size  $N \times 1$ ($N$ must be a power of two, i.e., $N = 2^J$, $J \in \mathbb{Z}^+$). The DWT of $Y$, denoted as  $d$, is represented as
\begin{equation}\label{eq-12}
  d = WY,
\end{equation}
where $W$ is an orthogonal matrix of size ${N \times N}$ the elements in $W$ are determined by the selection of a particular wavelet basis, such as Haar, Daubechies, and Symmlet. 

When $N$ is large, a fast algorithm based on filtering proposed by Mallat is used for computational efficiency. The DWT with this algorithm is obtained by performing of a series of successive convolutions that involve a wavelet-specific low-pass filter $h$ and its mirror counterpart, high-pass filter $g$. These repeated convolutions using the two filters accompanied with the operation of decimation (keeping every second coefficient of the convolution) generate a multiresolution representation of a signal, consisting of a smooth approximation ($c$-coefficients) and a hierarchy of detail coefficients $d_{jk}$ at different resolutions (denoted by a scale index $j$) and different locations within the same resolution, denoted by $k$. The convolutions with filters $h$ and $g$ are repeated until a desired decomposition level $j=J_0$ is reached ($1 \leq J_0 \leq J-1 $ and $J=\log_2 N$).

Thus, the vector  $\uw{d}$ in (\ref{eq-12}) has a structure,
\begin{equation}\label{eq-12a}
  \uw{d} = (\uw{c}_{J_0}, \uw{d}_{J_0}, \dots, \uw{d}_{J-2}, \uw{d}_{J-1}),
\end{equation}
where $\uw{c}_{J_0}$ is a vector of coefficient corresponding to a smooth trend in signal, and $\uw{d}_{j}$ are detail coefficients at different resolutions $j$ where $J_0 \leq j \leq J-1.$  It is the logarithm of the variance of coefficients in $\uw{d}_j$ that is used for definition of wavelet spectra.

\section{Standard Wavelet Spectra}\label{sec:appb}
Supposes  $\uw{d} = \{d_{1}, d_{2}, \cdots, d_{n}\}$ represents detail wavelet coefficients in the $j$th multiresolution level of wavelet transform of signal $Y$.  Then, the wavelet spectra of $Y$  is computed as follows.
\begin{equation}\label{WT_spectra}
    S(j) = \log_2(\hat{\sigma}^2(\uw{d}_j)), ~J_0 \leq j \leq J-1,
\end{equation}
where $\hat{\sigma}^2(\uw{d}_j)$ is an estimator of variance of vector $\uw{d}_j$.


\bibliographystyle{plain} 
\bibliography{sample}

\begin{thebibliography}{10}

\bibitem{R15}
Arin Chaudhuri and Wenhao Hu.
\newblock A fast algorithm for computing distance correlation.
\newblock {\em Comput. Stat. Data Anal.}, 135:15–24, jul 2019.

\bibitem{R32}
Benjamin Cowley, Joao Semedo, Amin Zandvakili, Matthew Smith, Adam Kohn, and
  Byron Yu.
\newblock {Distance Covariance Analysis}.
\newblock In Aarti Singh and Jerry Zhu, editors, {\em Proceedings of the 20th
  International Conference on Artificial Intelligence and Statistics},
  volume~54, pages 242--251. PMLR, 2017.

\bibitem{R16}
Dominic Edelmann, Donald Richards, and Daniel Vogel.
\newblock The distance standard deviation.
\newblock \url{https://arxiv.org/abs/1705.05777}, 2017.

\bibitem{R38}
Eric~T. Fung, Tai-Tung Yip, Lee Lomas, Zheng Wang, Christine Yip, Xiao-Ying
  Meng, Shanhua Lin, Fujun Zhang, Zhen Zhang, Daniel~W. Chan, and Scot~R.
  Weinberger.
\newblock Classification of cancer types by measuring variants of host response
  proteins using seldi serum assays.
\newblock {\em International Journal of Cancer}, 115(5):783--789, 2005.

\bibitem{R23}
Xiaoming Huo and Gabor~J. Sz{\'e}kely.
\newblock Fast computing for distance covariance.
\newblock \url{https://arxiv.org/abs/1410.1503}, 2014.

\bibitem{R4}
Seonghye Jeon, Orietta Nicolis, and Brani Vidakovic.
\newblock Mammogram diagnostics via 2-d complex wavelet-based self-similarity
  measures.
\newblock {\em São Paulo Journal of Mathematical Sciences}, 8(2):265--284,
  2014.

\bibitem{R10}
Yoon~Young Jung, Youngja Park, Dean~P. Jones, Thomas~R. Ziegler, and Brani
  Vidakovic.
\newblock Self-similarity in {NMR} spectra: An application in assessing the
  level of cysteine.
\newblock {\em Journal of Data Science}, 8(1):1--19, jul 2021.

\bibitem{R35}
Taewoon Kong and Brani Vidakovic.
\newblock Non-decimated complex wavelet spectral tools with applications.
\newblock \url{https://arxiv.org/abs/1902.01032}, 2019.

\bibitem{R3}
Lihua Li, Hong Tang, Zuobao Wu, Jianli Gong, Michael Gruidl, Jun Zou, Melvyn
  Tockman, and Robert~A Clark.
\newblock Data mining techniques for cancer detection using serum proteomic
  profiling.
\newblock {\em Artif Intell Med}, 32(2):71--83, 2004.

\bibitem{R27}
Runze Li, Wei Zhong, and Liping Zhu.
\newblock Feature screening via distance correlation learning.
\newblock {\em Journal of the American Statistical Association},
  107(499):1129--1139, 2012.

\bibitem{R31}
David~S. Matteson and Ruey~S. Tsay.
\newblock Independent component analysis via distance covariance.
\newblock {\em Journal of the American Statistical Association}, 112:623 --
  637, 2013.

\bibitem{R44}
Mary~E. McNamara, Mackenzie Zisser, Christopher~G. Beevers, and Jason Shumake.
\newblock Not just ``big'' data: Importance of sample size, measurement error,
  and uninformative predictors for developing prognostic models for digital
  interventions.
\newblock {\em Behaviour Research and Therapy}, 153:104086, 2022.

\bibitem{R14}
Pedro~Alberto Morettin.
\newblock {\em Wavelets in Statistics}.
\newblock University of S{\'a}o Paulo, S{\'a}o Paulo, Brazil, 1997.

\bibitem{R40}
Tiwari P, S~Viswanath, J~Kurhanewicz, A~Sridhar, and Madabhushi A.
\newblock Multimodal wavelet embedding representation for data combination
  (maweric): integrating magnetic resonance imaging and spectroscopy for
  prostate cancer detection.
\newblock {\em NMR Biomed}, 4(25):607--619, 2012.

\bibitem{R2}
Emanuel~F Petricoin, Ali~M Ardekani, Peter J~Levine Ben A~Hitt, Vincent~A
  Fusaro, Seth~M Steinberg, Gordon~B Mills, Charles Simone, David~A Fishman,
  Elise~C Kohn, and Lance~A Liotta.
\newblock Use of proteomic patterns in serum to identify ovarian cancer.
\newblock {\em Lancet}, 359(9306):572--577, 2002.

\bibitem{R39}
American National Cancer Institute~Internet Repository.
\newblock The datasets used in the study.
\newblock \url{https://home.ccr.cancer.gov/ncifdaproteomics/ppatterns.asp},
  2022.
\newblock Online; accessed on 05. 06.2022.

\bibitem{R34}
T~Roberts, Mary~S. Newell, William~F. Auffermann, and Brani Vidakovic.
\newblock Wavelet‐based scaling indices for breast cancer diagnostics.
\newblock {\em Statistics in Medicine}, 36:1989 -- 2000, 2017.

\bibitem{R8}
American~Cancer Society.
\newblock Key statistics for ovarian cancer.
\newblock
  \url{https://www.cancer.org/cancer/ovarian-cancer/about/key-statistics.html},
  2022.
\newblock Online; accessed July 7,2022.

\bibitem{R2a}
James~M Sorace and Min Zhan.
\newblock A data review and re-assessment of ovarian cancer serum proteomic
  profiling.
\newblock {\em BMC Bioinformatics}, 4(24):572--577, 2003.

\bibitem{R43}
Shonosuke Sugasawa and Genya Kobayashi.
\newblock Robust fitting of mixture models using weighted complete estimating
  equations.
\newblock {\em Computational Statistics \& Data Analysis}, 174:107526, 2022.

\bibitem{R36}
Liping Sun, Li~Li, Zhao Li, Shuhui Hong, Qifeng Yang, Xun Qu, and Beihua Kong.
\newblock Alterations in the serum proteome profile during the development of
  ovarian cancer.
\newblock {\em International journal of oncology}, 45-6:2495--501, 2014.

\bibitem{R25}
G{\'a}bor~J. Sz{\'e}kely and Maria~L. Rizzo.
\newblock {Brownian distance covariance}.
\newblock {\em The Annals of Applied Statistics}, 3(4):1236 -- 1265, 2009.

\bibitem{R24}
G{\'a}bor~J. Sz{\'e}kely and Maria~L. Rizzo.
\newblock On the uniqueness of distance covariance.
\newblock {\em Statistics \& Probability Letters}, 82(12):2278--2282, 2012.

\bibitem{R26}
Gabor~J. Sz{\'e}kely and Maria~L. Rizzo.
\newblock Partial distance correlation with methods for dissimilarities.
\newblock \url{https://arxiv.org/abs/1310.2926}, 2013.

\bibitem{R17}
Hong Tang, Y.~Mukomel, and E.~Fink.
\newblock Diagnosis of ovarian cancer based on mass spectra of blood samples.
\newblock In {\em 2004 IEEE International Conference on Systems, Man and
  Cybernetics (IEEE Cat. No.04CH37583)}, volume~4, pages 3444--3450. IEE, 2004.

\bibitem{R7}
Lindsey~A. Torre, Britton Trabert, Carol~E. DeSantis, Kimberly~D. Miller, Goli
  Samimi, Carolyn~D. Runowicz, Mia~M. Gaudet, Ahmedin Jemal, and Rebecca~L.
  Siegel.
\newblock Ovarian cancer statistics, 2018.
\newblock {\em CA: A Cancer Journal for Clinicians}, 68(4):284--296, 2018.

\bibitem{R33}
Emmanuel~Selorm Tsyawo and Abdul-Nasah Soale.
\newblock A distance covariance-based estimator.
\newblock \url{https://arxiv.org/abs/2102.07008}, 2021.

\bibitem{R42}
Khalid Usman and Kashif Rajpoot.
\newblock Brain tumor classification from multi-modality mri using wavelets and
  machine learning.
\newblock {\em Pattern Analysis and Applications}, 20(3):871--881, 2017.

\bibitem{R1}
Marina Vannucci, Naijun Sha, and Philip~J. Brown.
\newblock Nir and mass spectra classification: Bayesian methods for
  wavelet-based feature selection.
\newblock {\em Chemometrics and Intelligent Laboratory Systems},
  77(1):139--148, 2005.

\bibitem{R21}
Brani Vidakovic.
\newblock {\em Statistical modeling by wavelets}.
\newblock John Wiley and Sons Inc, New York, NY, USA, 1999.

\bibitem{R18}
Brani Vidakovic.
\newblock {\em Engineering Biostatistics : An Introduction Using Matlab and
  Winbugs. INSERT-MISSING-SERVICE-NAME}.
\newblock Hoboken, New Jersey: John Wiley \& Sons, 2007.

\bibitem{R41}
Fei Wang, Xian-Hua Han, and Yen-Wei Chen.
\newblock Biomedical imaging modality classification using combined visual
  features and textual terms.
\newblock {\em International Journal of Biomedical Imaging}, 2011:241396, 2011.

\bibitem{R37}
Hui Zhang, Beihua Kong, Xun Qu, Lin Jia, Biping Deng, and Qifeng Yang.
\newblock Biomarker discovery for ovarian cancer using seldi-tof-ms.
\newblock {\em Gynecologic Oncology}, 102(1):61--66, 2006.

\end{thebibliography}

\end{document}